# Enhancing Person Re-Identification through Tensor Feature Fusion

*Akram Abderraouf Gharbi[1], Ammar Chouchane[2], Mohcene Bessaoudi[3], Abdelmalik Ouamane[4], El ouanas Belabbaci[5]

[1] University of Biskra, Algeria, akram.gharbi@univ-biskra.dz

[2] University Center of Barika, Algeria, chouchane_ammar@yahoo.com

[3] University of Biskra, Algeria, bessaoudi.mohcene@gmail.com

[4] University of Biskra, Algeria, ouamane.abdelmalik@univ-biskra.dz

[5] University of Bejaia, Algeria, elouanas.belabbaci@univ-bejaia.dz

**Abstract**

In this paper, we present a novel person reidentification (PRe-ID) system that based on tensor feature representation and multilinear subspace learning. Our approach utilizes pretrained CNNs for high-level feature extraction, along with Local Maximal Occurrence (LOMO) and Gaussian Of Gaussian (GOG ) descriptors. Additionally, Cross-View Quadratic Discriminant Analysis (TXQDA) algorithm is used for multilinear subspace learning, which models the data in a tensor framework to enhance discriminative capabilities. Similarity measure based on Mahalanobis distance is used for matching between training and test pedestrian images. Experimental evaluations on VIPeR and PRID450s datasets demonstrate the effectiveness of our method.

**Introduction**

In the past few years, artificial intelligence has sparked a transformative revolution across multiple domains, significantly impacting people's lives. Among its many applications in areas such as security, commerce, healthcare, and education, smart video surveillance stands out [1]. By incorporating computer vision through machine learning techniques, smart video surveillance encompasses various tasks, including person reidentification. Person Re-Identification (PRe-ID) specifically aims to determine the similarity between images of individuals captured by non-overlapping cameras [2]. This task plays a crucial role in enhancing security and identification processes within the field of video surveillance. Firstly, the goal of person PRe-ID is detecting the similarity between person images sorted from non-overlapping cameras. Several problems are existing with PRe-ID systems due to various factors, such as low resolution, occlusion, illumination variation, and pose changes [3]. PRe-ID system contains three principal phases: features extraction, learning, and mathing scores. To achieve the best scores result many works propose different types of descriptors for robust data representation. Some of those descriptors focus on local features like color, texture, edges, and contours, Shengcai Liao et.al [4] proposed Local Maximal Occurrence (LOMO) to process viewpoint changes and illumination variations. Tetsu Matsukawa et al. [5] applied a Hierarchical Gaussian descriptor (GOG) on each local region of the image to extract local features. In [6] they proposed a global representation of an image through LDFV descriptor, which represents local features like image intensity, and color. Also in [7] they used a Hybrid Spatiogram which was collected from several color channels to extract the features. Some others use deep learning and transfer learning models to extract the deep features [8]–[11], these approaches aim to enhance feature representation to improve the performances. Another side, some works propose methods to improve the learning stage like XQDA [4], EquiDML [12], and KRKISS [13] by increasing the between class and decreasing the within class.

Our contributions in this paper are summarized in the following:

---

Akram Abderraouf Gharbi[1]

1. We propose a new multilinear representation of features based on pretrained Convolutional Neural Networks (CNNs) as a high-level feature extractor, merged with two local features descriptors Local Maximal Occurrence (LOMO) and Gaussian Of Gaussian (GOG). The fusion of these features in a tensor representation allows for capturing both spatial and semantic information, leading to improved discriminative capabilities.

2. Additionally, we propose a new multilinear subspace projection algorithm named (Tensor-based Cross-View Quadratic Discriminant Analysis). TXQDA effectively exploits the inter-camera correlations and inter-modal interactions present in the tensor feature representation. By modeling the data in a multilinear framework, it learns a discriminative subspace that enhances the separability between different individuals.

3. To evaluate the proposed method, we conducted experiments on two challenging person re-identification datasets, VIPeR and PRID450s. The matching-based Mahalanobis distance is employed for similarity computation between query and gallery samples. The experimental results demonstrate the effectiveness of our approach, achieving significantly improved re-identification performance compared to existing methods. The obtained results affirm the potential of tensor-based fusion and multilinear subspace learning techniques in advancing the field of person re-identification.

The rest of this paper is organized as follow, the proposed methodology is provided in Section II, in which we describe the Tensor feature representation scheme and the Multilinear subspace learning process. Then, the Results and discussions are presented in Section III. Finaly, conclusion and future directions are given in Section IV.

**Methodology**

**A. Proposed Person Re-Identification approach** The proposed approach of Person Re-identification (PReID) system is shown in Fig. 1 Generally, the mechanism of Person Re-identification is to detect similar images of the candidate person through the gallery image datasets of various cameras. So, the system ranks the images of persons relying on the similarity with the probe. Based on the previous, our system includes of three essential stages: Design the Tensor of the extracted features, multilinear subspace learning, and matching with score normalization.

**B. Tensor feature representation and multilinear subspace learning**

To extract the image features, Three descriptors are used CNN [14] for deep features, LOMO [4] and GOG [5] for shallow features to produce three features vectors for each person image of the gallery, these descriptors are very effective in low- resolution, lighting, viewpoint, and background variations. For robust representation, each vector splits into parts to create a 3-order tensor, each tensor has three modes, the first one represents the number of feature parts, the second mode is the features, and the third mode represents the persons. Then, we combine the tensors to get CNN+LOMO Tensor and CNN+GOG Tensor. Fig. 2 Illustrates the technique of feature extraction and tensor design.

In the offline training phase, the proposed technique TXQDA projects the training tensors X and Y on a new discriminant subspace, and the dimensions of both tensors are reduced to obtain new dimensions $l'_1 \times l'_2$ for mode-1 and mode-2 respectively, where $l'_1 \times l'_2 \ll l_1 \times l_2$. The dimension of mode-3 represents the persons in the database, so it remains the same. Fig. 3 depicts the different steps of TXQDA algorithm. In the online testing phase, the same procedures occur on each probe of pair images. After the projection, the matching of a probe with the gallery is performed by computing the Mahalanobis distance [15] in the new discriminant subspace.

**Results and discussion**

**A. Dataset**

We evaluated the effectiveness of our approach by comprised of comparing our experiment against state-of-the-art methods, on two challenging PReID datasets: VIPeR [16], and PRID450S [15].

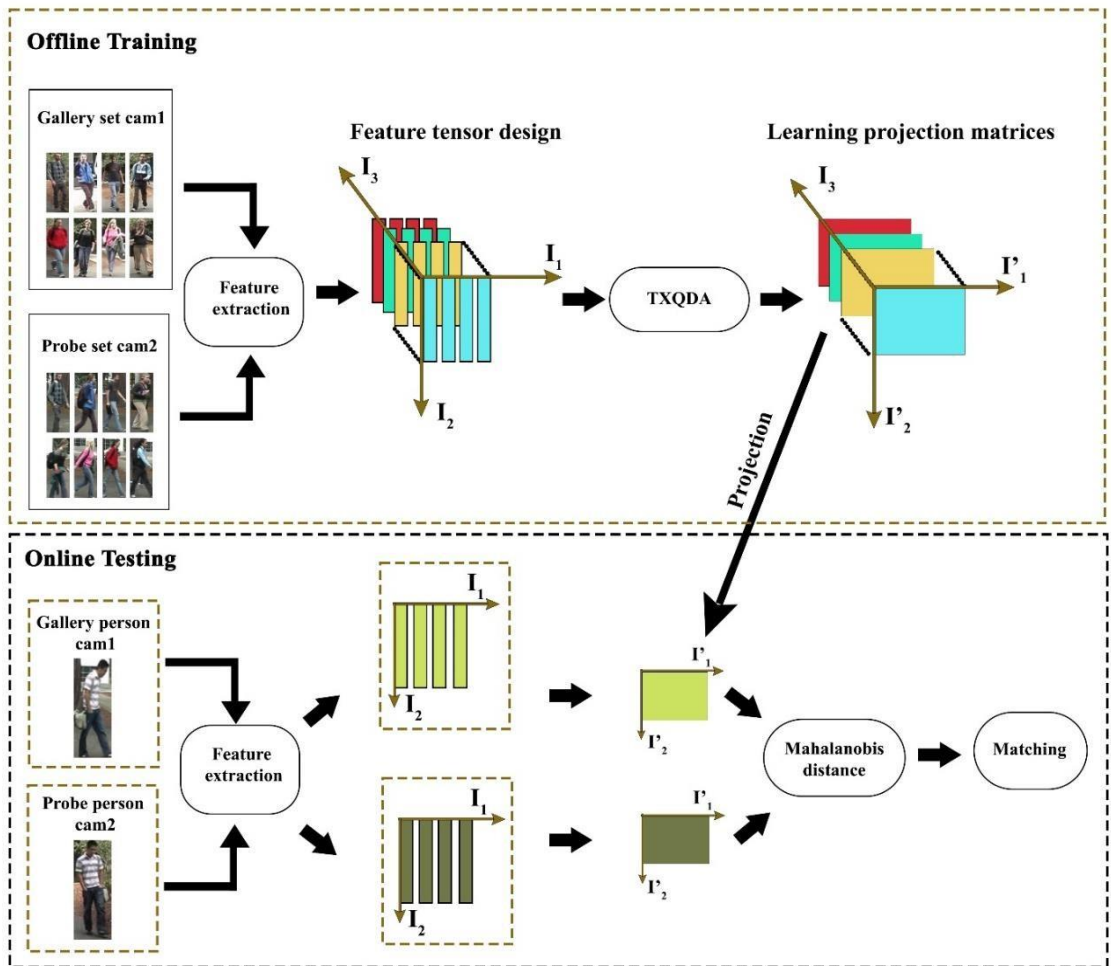

Fig. 1 The proposed approach pipeline of Person Re-id

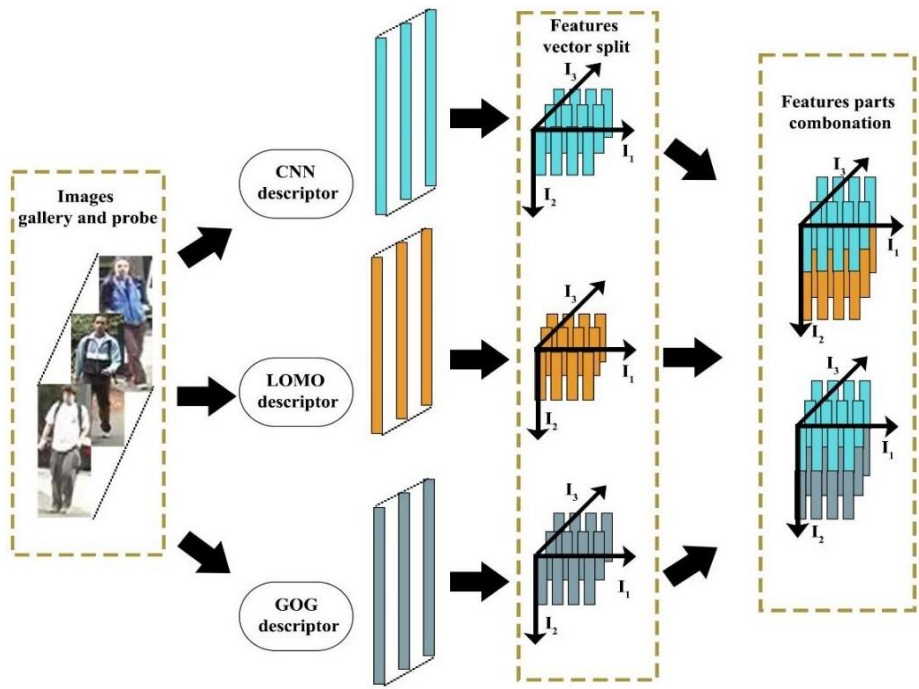

Fig. 2 The tensors design process

### 1) ViPeR Dataset

It captured from two surveillance cameras and contains 632 pairs of person images. Each image in this dataset have a size 128x48 pixels. The main challenges in this dataset are variations in illumination, and viewpoint.

### 2) PRID450S Dataset

It captured from two cameras and contains 450 pair of identitie images. Each image in this dataset have a size 128x48 pixels. Images in this dataset contain illumination and pose variation, significant differences in background and viewpoint.

We use the 10-folder cross-validation protocol [17] in the experiments and evaluate the performance of our approach using the cumulative matching characteristics (CMC) curves. The results into rank-1, rank-5, rank-10, rank-15, and rank-20 are given. The experimental results of the two used datasets ViPeR, PRID450s are shown in Table I and Table II respectively, and their CMC curves are illustrated in Fig. 4 and Fig. 5.

### B. Tensors features comparison

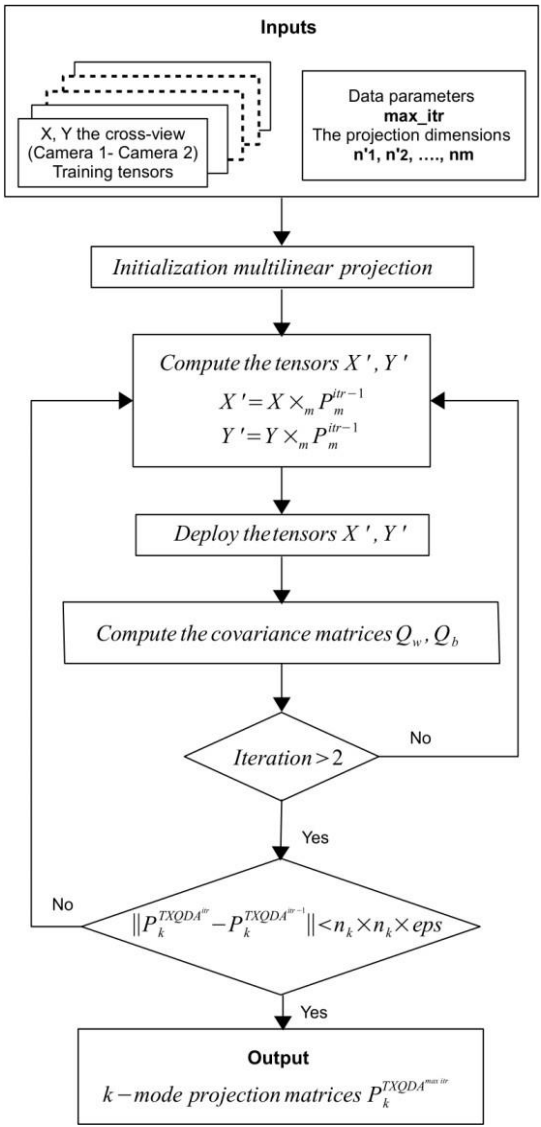

Fig. 3 TXQDA algorithm

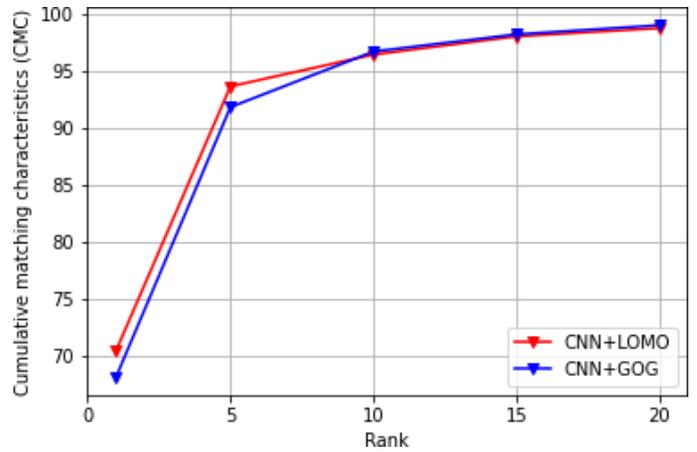

Fig. 4 CMC curves of the best features on PRID450s database

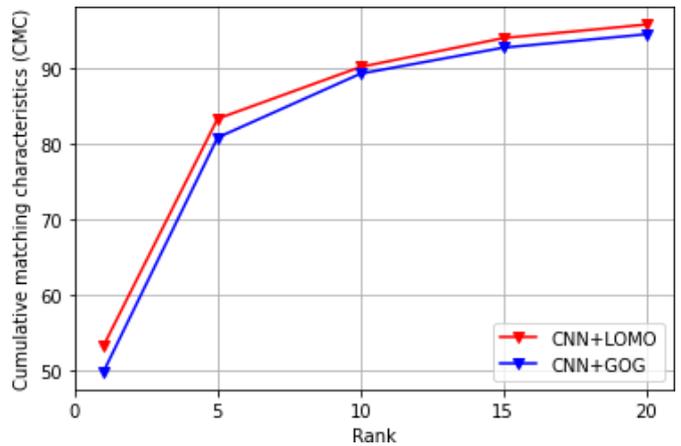

Fig. 5 CMC curves of the best features on ViPeR database

Generally, CNN+LOMO tensor has achieved the best results against CNN+GOG tensor, but the TXQDA output dimension has a role in the variation of the scores percentages. In PRID450s dataset, the rank-1 of CNN+LOMO tensor reached 70.40% in dimension 200 and decreased in dimension 250, while the rank-1 of CNN+GOG tensor reached its best result 68.09% in dimension 250. In ViPeR dataset, the two tensors reached the best result in dimension 250 where CNN+LOMO achieved 53.16% and CNN+GOG achieved 49.72%.

**C. Comparison against the state-of-the-art**

We have compared our method with previous work methods in recent years, and have achieved the best scores results in the used datasets. Table III shows our technique against the state-of-the-art works from rank-1 to rank-20.

Table I: Performance analysis and results on prid450s

| Features types | Dim | Matching rates (%) | | | |
|---|---|---|---|---|---|
| | | **Rank-1** | **Rank-5** | **Rank-10** | **Rank-20** |
| **CNN+LOMO** | 50 | 62.53 | 90.09 | 95.38 | 98.27 |
| | 100 | 67.73 | 92.76 | 96.53 | 98.80 |
| | 150 | 69.82 | 93.38 | 96.40 | 98.80 |
| | 200 | 70.40 | 93.64 | 96.44 | 98.76 |
| | 250 | 69.78 | 93.02 | 96.22 | 98.67 |
| **CNN+GOG** | 50 | 57.73 | 87.07 | 94.00 | 98.00 |
| | 100 | 64.80 | 90.49 | 96.04 | 98.76 |
| | 150 | 66.18 | 91.42 | 96.36 | 98.84 |
| | 200 | 67.20 | 91.91 | 96.67 | 98.93 |
| | 250 | 68.09 | 91.82 | 96.71 | 99.02 |

Table II: Performance analysis and results on viper

| Features types | Dim | Matching rates (%) | | | |
|---|---|---|---|---|---|
| | | **Rank-1** | **Rank-5** | **Rank-10** | **Rank-20** |
| **CNN+LOMO** | 50 | 62.53 | 90.09 | 95.38 | 98.27 |
| | 100 | 67.73 | 92.76 | 96.53 | 98.80 |
| | 150 | 69.82 | 93.38 | 96.40 | 98.80 |
| | 200 | 70.40 | 93.64 | 96.44 | 98.76 |
| | 250 | 69.78 | 93.02 | 96.22 | 98.67 |
| **CNN+GOG** | 50 | 57.73 | 87.07 | 94.00 | 98.00 |
| | 100 | 64.80 | 90.49 | 96.04 | 98.76 |
| | 150 | 66.18 | 91.42 | 96.36 | 98.84 |
| | 200 | 67.20 | 91.91 | 96.67 | 98.93 |
| | 250 | 68.09 | 91.82 | 96.71 | 99.02 |

Table III: Comparison with the sota of rank-1 and rank-20 identification rates (%) on the viper and prid450s datasets.

| Approach | Year | VIPeR | | PRID450s | |
|---|---|---|---|---|---|
| | | Rank-1 | Rank-20 | Rank-1 | Rank-20 |
| FT-CNN+XQDA [8] | 2016 | 42.50 | 92.00 | 58.20 | 94.30 |
| SSDAL+XQDA [18] | 2016 | 43.50 | 89.00 | 22.60 | 69.20 |
| Kernel X-CRC [19] | 2019 | 51.60 | 95.30 | 68.80 | 98.40 |
| VS-SSL [20] | 2020 | 43.90 | 87.80 | 63.30 | 97.00 |
| SLDL [21] | 2022 | 51.23 | 95.02 | - | - |
| CNN+LOMO+TXQDA(Our) | 2022 | 53.16 | 95.82 | 70.40 | 98.76 |
| CNN+GOG+TXQDA(Our) | 2023 | 49.72 | 94.53 | 68.09 | 99.02 |

**Conclusion**

In this paper, we propose a novel approach for person reidentification by combining tensor feature representation and multilinear subspace learning techniques. Our method leverages the power of pretrained CNN as a high-level feature extractor, along with two complementary descriptors, LOMO and GOG. Furthermore, we incorporate a new multilinear subspace projection named TXQDA which is used to reduce the high dimension of the high-order tensor and create a new features representation with augmentation of the inter-class and detraction of the intra-class. As a future work, we propose to develop a new deep model trained by all the available datasets of Person re-identification in order to improve our system.

**References**

[1] V. D. Păvăloaia and S. C. Necula, "Artificial Intelligence as a Disruptive Technology—A Systematic Literature Review," *Electron.*, vol. 12, no. 5, 2023, doi: 10.3390/electronics12051102.
[2] C. Ammar, B. Mohcene, B. Elhocine, and A. Ouamane, "A new multidimensional discriminant representation for robust person re-identification," *Pattern Anal. Appl.*, no. 0123456789, 2023, doi: 10.1007/s10044-023-01144-0.
[3] G. Zou, G. Fu, X. Peng, Y. Liu, M. Gao, and Z. Liu, "Person re-identification based on metric learning: a survey," *Springer*, pp. 26855–26888, 2021.
[4] S. Liao, Y. Hu, X. Zhu, and S. Z. Li, "Person re-identification by local maximal occurrence representation and metric learning," in *Proceedings of the IEEE conference on computer vision and pattern recognition*, 2015, pp. 2197–2206.
[5] T. Matsukawa, T. Okabe, E. Suzuki, and Y. Sato, "Hierarchical Gaussian Descriptor for Person Re-identification," *Proc. IEEE Comput. Soc. Conf. Comput. Vis. Pattern Recognit.*, vol. 2016-Decem, pp. 1363–1372, 2016, doi: 10.1109/CVPR.2016.152.
[6] B. Ma, Y. Su, and F. Jurie, "Local descriptors encoded by fisher vectors for person re-identification," in *European conference on computer vision*, Springer, 2012, pp. 413–422.
[7] Z. Mingyong, Z. Wu, C. Tian, Z. Lei, and H. Lei, "Efficient person re-identification by hybrid spatiogram and covariance descriptor," *IEEE Comput. Soc. Conf. Comput. Vis. Pattern Recognit. Work.*, vol. 2015-Octob, pp. 48–56, 2015, doi: 10.1109/CVPRW.2015.7301296.
[8] T. Matsukawa and E. Suzuki, "Person re-identification using CNN features learned from combination of attributes," *Proc. - Int. Conf. Pattern Recognit.*, vol. 0, pp. 2428–2433, 2016, doi: 10.1109/ICPR.2016.7900000.
[9] M. Niall, M. del R. Jesus, and M. Paul, "Recurrent_Convolutional_Network_CVPR_2016_paper," *Comput. Vis. Found.*, vol. 9905, no. i, pp. 1–11, 2016, doi: 10.1017/CBO9781107415324.004.
[10] H. Wang and J. Hu, "Deep Multi-Task Transfer Network for Cross Domain Person Re-Identification," *IEEE Access*, vol. 8, pp. 5339–5348, 2020, doi: 10.1109/ACCESS.2019.2962581.
[11] A. Gupta, P. Pawade, and R. Balakrishnan, "Deep Residual Network and Transfer Learning-based Person Re-Identification," *Intell. Syst. with Appl.*, vol. 16, no. April, p. 200137, 2022, doi: 10.1016/j.iswa.2022.200137.

[12] J. Wang, Z. Wang, C. Liang, C. Gao, and N. Sang, "Equidistance constrained metric learning for person re-identification," *Pattern Recognit.*, vol. 74, pp. 38–51, 2018, doi: 10.1016/j.patcog.2017.09.014.

[13] C. Zhao, Y. Chen, X. Wang, W. K. Wong, D. Miao, and J. Lei, "Kernelized random KISS metric learning for person re-identification," *Neurocomputing*, vol. 275, pp. 403–417, 2018, doi: 10.1016/j.neucom.2017.08.064.

[14] A. Krizhevsky, I. Sutskever, and G. E. Hinton, "ImageNet classification with deep convolutional neural networks," *Commun. ACM*, vol. 60, no. 6, pp. 84–90, 2017, doi: 10.1145/3065386.

[15] P. M. Roth, M. Hirzer, M. Köstinger, C. Beleznai, and H. Bischof, "Mahalanobis distance learning for person re-identification," *Pers. re-identification*, pp. 247–267, 2014.

[16] D. Gray and H. Tao, "Viewpoint invariant pedestrian recognition with an ensemble of localized features," *Lect. Notes Comput. Sci. (including Subser. Lect. Notes Artif. Intell. Lect. Notes Bioinformatics)*, vol. 5302 LNCS, no. PART 1, pp. 262–275, 2008, doi: 10.1007/978-3-540-88682-2_21.

[17] J. D. Rodríguez, A. Pérez, and J. A. Lozano, "Sensitivity Analysis of k-Fold Cross Validation in Prediction Error Estimation," *IEEE Trans. Pattern Anal. Mach. Intell.*, vol. 32, no. 3, pp. 569–575, 2010, doi: 10.1109/TPAMI.2009.187.

[18] C. Su, S. Z. B, J. Xing, W. Gao, and Q. Tian, "Deep Attributes Driven Multi-camera Person," *Eccv*, vol. 1, pp. 475–491, 2016, doi: 10.1007/978-3-319-46475-6.

[19] J. Jia, Q. Ruan, Y. Jin, G. An, and S. Ge, "View-specific subspace learning and re-ranking for semi-supervised person re-identification," *Pattern Recognit.*, vol. 108, 2020, doi: 10.1016/j.patcog.2020.107568.

[20] G. Zhang, T. Jiang, J. Yang, J. Xu, and Y. Zheng, "Cross-view kernel collaborative representation classification for person re-identification," *Multimed. Tools Appl.*, vol. 80, no. 13, pp. 20687–20705, 2021, doi: 10.1007/s11042-021-10671-z.

[21] J. Sun, L. Kong, and B. Qu, "Sparse and Low-Rank Joint Dictionary Learning for Person Re-Identification," *Mathematics*, vol. 10, no. 3, pp. 1–16, 2022, doi: 10.3390/math10030510.